\documentclass{article}
\usepackage{arxiv}

\usepackage[utf8]{inputenc} 
\usepackage[T1]{fontenc}    
\usepackage{hyperref}       
\usepackage{url}            
\usepackage{booktabs}       
\usepackage{amsfonts}       
\usepackage{nicefrac}       
\usepackage{microtype}      
\usepackage{lipsum}
\usepackage{graphicx}
\usepackage{amsmath} 

\usepackage{blindtext,alltt}

\usepackage{longtable}

\title{Exploring the encoding of linguistic representations in the Fully-Connected Layer of generative CNNs for Speech}

\author{
Bruno Ferenc \v{S}egedin \\
  Program in Linguistics\\
  Brown University\\
  Providence, RI 02912 \\
  \texttt{bruno\_ferenc\_segedin@brown.edu} \\
   \And
 Ga\v{s}per Begu\v{s} \\
  Department of Linguistics\\
  University of California, Berkeley\\
  Berkeley, CA 94720 \\
  \texttt{begus@berkeley.edu} \\
  }

\begin{document}

\maketitle

\begin{abstract}
Interpretability work on the convolutional layers of CNNs has primarily focused on computer vision, but some studies also explore correspondences between the latent space and the output in the audio domain. However, it has not been thoroughly examined how acoustic and linguistic information is represented in the fully-connected (FC) layer that bridges the latent space and convolutional layers. The current study presents the first exploration of how the FC layer of CNN’s for speech synthesis encodes linguistically relevant information. We propose two techniques for exploration of the fully-connected layer. In Experiment 1, we use weight matrices as inputs into convolutional layers. In Experiment 2, we manipulate the FC layer to explore how symbolic-like representations are encoded in CNNs. We leverage the fact that the FC-layer outputs a feature map and that variable-specific weight matrices are temporally structured to (i) demonstrate how the distribution of learned weights varies between latent variables in systematic ways and (ii) demonstrate how manipulating the FC layer while holding constant subsequent model parameters affects the output. We ultimately present an FC manipulation that can output a single segment. Using this technique, we show that lexically-specific latent codes in generative CNNs (ciwGAN) have shared lexically-invariant sublexical representations in the FC-layer weights, showing that ciwGAN encodes lexical information in a linguistically-principled manner.  

\end{abstract}

\section{Introduction}
\subsection{Overview}
This study explores how linguistically relevant information is represented in the fully-connected layer of generative CNNs. We focus on ciwGAN (\cite{beguvs2021ciwgan}), a convolutional neural network (CNN) in the Generative Adversarial Network framework (\cite{goodfellow2020generative}) that is optimized to generate human-like speech from a compact set of latent variables, while also being optimized to produce lexically informative outputs. Following prior work (e.g. \cite{beguvs2020modeing},\cite{beguvs2021ciwgan}, \cite{chen2023exploring}), we investigate how variability in the models' outputs is mapped to its latent space. The current study departs from prior work in that we directly analyze the learned weights of the fully-connected layer that bridges the latent space and subsequent convolutional layers.

\subsection{Background}
To acquire a spoken language, humans must learn to compress variable, non-linear, and stochastic input into a finite set of symbolic or discrete representations that they can recombine compositionally. How humans accomplish this task, and the requisite knowledge and cognitive capacities necessary for doing so, are still a subject of much investigation (\cite{baroni2020linguistic}, \cite{belinkov2017analyzing}, \cite{feldman2009influence}, \cite{barreda2020vowel}, \cite{raviv2018systematicity}, \cite{saffran1999statistical}, \cite{grieser1989categorization}). Deep neural networks (DNNs) offer a framework for modeling this learning task and its consequences on speech production and perception. While various linear dimensionality reduction techniques exist, such as principal components analysis (PCA), DNNs are uniquely equipped for modeling the compression of the kind of highly variable and non-linear data that is characteristic of human speech (e.g. \cite{shain-elsner-2020-acquiring}, \cite{beguvs2020modeing}). 

One framework for modeling the learning of representations using DNNs are latent space models. While latent space models have been frequently used for for image and audio generation (e.g. \cite{wang2020cnn}, \cite{taigman2016unsupervised}), only recently have researchers begun to use them for cognitive modeling of phonetic and phonological learning (\cite{barman2024unsupervised}, \cite{beguvs2021identity},\cite{beguvs2023articulation},\cite{beguvs2021identity}, \cite{shain2019measuring}). One reason latent space models are appealing when it comes to linguistic and cognitive modeling is that they explicitly model the compression of highly variable and complex continuous data, such as waveforms of lexical items, into a compact set of variables or codes (e.g. \cite{beguvs2020modeing}, \cite{shain2019measuring}, \cite{shain2019measuring}). The training objective of these models is, in broad terms, for the model's output to resemble its training inputs. To achieve this objective with the constraint of a compact latent space, the model must encode output variability in the latent space in a principled or generalizable way, offering a test of its capacity to learn theory-driven linguistic generalizations. Ultimately, if the coding scheme in the latent space that emerges from this task reflects aspects human linguistic knowledge, such an outcome can indicate the trained network is a plausible model of human phonological acquisition (e.g. \cite{beguvs2021ciwgan}), or that the conditions necessary for constructing such generalization is available in the learning signal (e.g. \cite{shain2019measuring}).

\subsection{Generative Adversarial Models of Language Acquisition}

Recent work has evaluated Generative Adversarial networks (GANs) as models of phonological learning and acquisition (\cite{beguvs2021identity}). GANs are characterized by a zero-sum competition between two networks: a generator whose task is to generate waveform outputs, and a discriminator whose task is to discriminate real training data from data produced by the generator. Critically, the generator is optimized to fool the discriminator, and is thus indirectly optimized to produce outputs that resemble the training data. One advantage of this framework for modeling the acquisition of phonetic and phonological patterns in speech is that the generator does not have access to the training data, and thus can make linguistically informative over-generalizations in its outputs. For example, it has been found that the GANs learn allophonic aspiration (\cite{beguvs2020modeing}), and that they learn the morpho-phonological process of reduplication and can generalize beyond their training data (\cite{beguvs2021identity}). In the latter study, a ciwGAN model was trained on reduplicated and unreduplicated stems, while a subset of stems did not have corresponding training items with a reduplicative prefix. After training, this model was capable of producing a reduplicative prefix for the stem that did not receive a prefix in training. These findings suggest that the pressure to map patterns to the latent space yields latent variables that appear to accomplish rule-like operations like copying, that manipulate symbolic-like linguistic categories. 

A pervasive challenge of any modeling approach that relies on DNNs is the "black box" problem: a model may exhibit some human-like behavior but the correspondence of the underlying states and processes to human behavior is unclear (e.g. \cite{abdullah2021familiar}, \cite{baas2024disentanglement}). For models like WaveGAN and ciw/fiwGAN, it is possible that while the model's outputs resemble training data, the model represents this knowledge in unsystematic or intuitively indecipherable ways. Some of the aforementioned work on the interpretability of GANs involves analyzing the internal states of the network and the relationships between internal representations. Specifically, this work has discovered that GANs encode different linguistic features at different convolutional layers in the generator and the Q-network (\cite{beguvs2022interpreting}, \cite{begusZhouIEEE}). Much of this work has contributed to the interpretability of GANs by demonstrating correspondences between latent variables and output values which shows that the latent space partitions output data into meaningful units or dimensions (e.g. \cite{beguvs2021ciwgan}, \cite{beguvs2021identity}, \cite{begusCSL} \cite{chen2023exploring}). In this paper, we present an approach for interpretability of generative CNNs that, to our knowledge, has not been used for models of speech: we draw inferences about the latent space representations by examining  variable-specific weight matrices and systematically manipulating sub-parts of the dense layer's output to investigate how linguistic information encoded within and across these variable-specific fully-connected layer encodings. 

\subsection{Current Study}

The current study shows how both the variable-specificity of the trained weight matrices and their feature-map structure can be leveraged to aid in model interpretability. While prior work has investigated the model's learning by manipulating the latent variables, including a handful of studies that have done so using GANs (e.g.~\cite{beguvs2021identity}), our study builds on interpretability of these models by bypassing latent space interpolation entirely and instead characterizing latent variables in terms of their weight matrices. 

In Experiment 1, we show that latent variables can be represented using variable-specific weight matrices that have a time-series feature map structure, and that comparing variables by their weight matrices can aid in bottom-up exploration of the latent space. We also demonstrate that weight matrices can be passed as inputs into the subsequent convolutional layers to produce interpretable waveform outputs. In Experiment 2, we show that sub-parts of variable-specific weight matrices can be extracted and used to generate waveforms in isolation. We also show that subparts of the FC layer can be predictably manipulated to output a phonological segment. We use this set of novel techniques to demonstrate that, when latent variables are mapped to lexical items, variable-specific weight matrices can contain evidence of lexically-invariant sublexical structure. This study thus contributes to the interpretability of ciwGAN by showing that rather than encoding lexical items wholistically in an indecipherable way, the model makes use of common sublexical structures across lexical representations.

\begin{figure}
    \centering
    \includegraphics[width=1\linewidth]{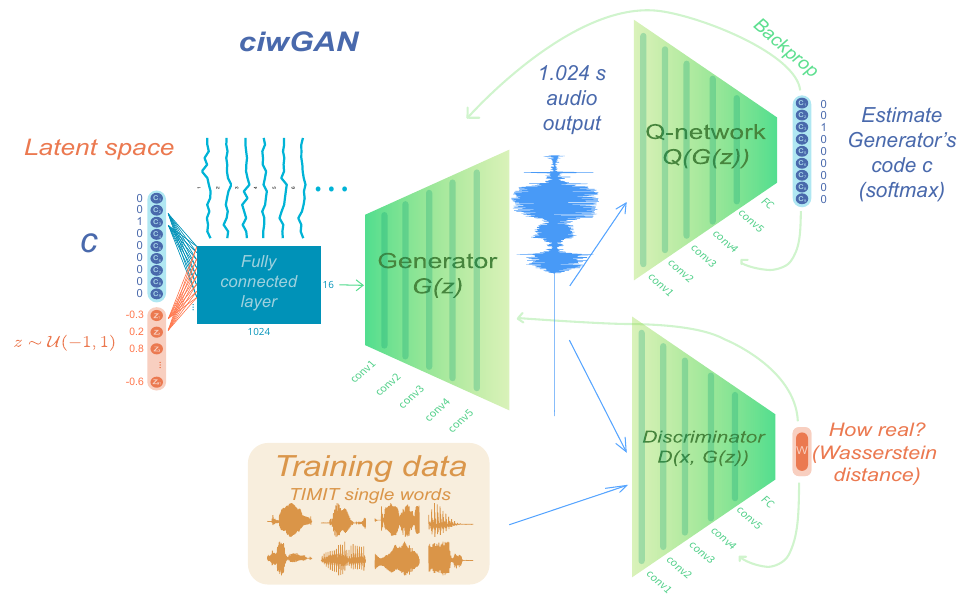}
    \caption{The model architecture of CiwGAN. The diagram is based on \cite{begus22Interspeech}. Throughout this paper, we refer to the axis represented vertically in the generator diagram as the ``time-axis''. }
    \label{fig:model_schematic}
\end{figure}

\section{Methods}
\subsection{Model \& Training}

The current study uses ciwGAN (\cite{beguvs2021ciwgan}), a version of  WaveGAN \cite{donahue2018adversarial}). The model architecture is shown in Fig.~\ref{fig:model_schematic}. The basic WaveGAN model learns to generate waveform outputs after being trained on raw audio data. Specifically, the training paradigm is characterized by a competition between two distinct networks: the generator and the discriminator. The generator samples from a latent space (\textit{z}) to produce waveform outputs, while the discriminator is optimized to distinguish between outputs produced by the generator and real training data (\textit{x}). Critically, the generator is optimized to fool the discriminator (equation 1), while the discriminator is optimized to classify training and fake data as accurately as possible.  More specifically, at each training step, the discriminator is optimized to maximize the Wasserstein distance between the scores it assigns a batch of generated outputs, and the scores it assigns a batch of training samples. 

\begin{equation}
    \mathcal{L}_G = -\mathbb{E}_{\mathbf{z} \sim p_{\mathbf{z}}(\mathbf{z})} \left[ D(G(\mathbf{z})) \right]
    \label{eq:wavegan_generator_loss}
\end{equation}
\begin{equation}
\mathcal{L}_D = \mathbb{E}_{\mathbf{x} \sim \mathbb{P}_r} [D(\mathbf{x})] - \mathbb{E}_{\mathbf{z} \sim \mathbb{P}_z} [D(G(\mathbf{z}))]
\label{eq:discriminator_loss_wavegan}
\end{equation}
 
CiwGAN is an innovation of WaveGAN which consist of a third network, the Q-network (based on InfoGAN \cite{chen16}). In ciwGAN, a subset of the latent variables that initialize the generator are implemented as a one hot encoding. The Q-network's task is to correctly estimate the one-hot code that initialized that generator's output. Both the generator and the Q-network are optimized to improve the Q-network's accuracy. In effect, this means that the generator is pressured to maximize the mutual information between latent codes and aspects of the output. This in turn manifests as a `communicative' pressure for the generator to not only produce well-formed outputs, but also to produce informative outputs (equation 3). The Q-network's loss function is also used to updated the weights of the generator. 

\begin{equation}
    \mathcal{L}_Q(\theta_Q) = \mathbb{E}_{s, a \sim \text{data}} \left[ (r(s, a) + \gamma \max_{a'} Q_{\theta_Q}(s', a') - Q_{\theta_Q}(s, a))^2 \right]
    \label{eq:ciwgan_q_loss}
\end{equation}

The current implementation of ciwGAN was trained on 5,803 tokens from the TIMIT corpus (\cite{garofolo1993timit}), that comprise the following 9 lexical items: "greasy", "ask", what, "water", carry", "year", "back", "like" and "suit". We test the model after 500 epochs (45000 training steps). Each model output is generated by a tensor of 91 uniformly distributed variables ($z \sim U(-1, 1)$), concatenated with a one-hot encoding tensor for 9 latent codes, which is the vector which the Q-network is optimized to classify. The generator is also optimized to help the Q-network classify this latent code by ensuring that the 9-value vector, after soft-max, reflects the one-hot encoding of the latent code that initializes the network. Prior work has demonstrated the success of ciwGAN in mapping N lexical types in the training data to the N distinct latent codes (\cite{beguvs2021ciwgan}), and the current implementation showed similar success, according to metrics used by prior studies. These results are not the subject of the analysis; we use the successful performance of the current model as the starting point for analyzing how lexical learning is reflected in the weights of the fully connected layer. 

\subsection{The fully-connected layer in generative latent space models}

\begin{figure}
    \centering
    \includegraphics[width=0.85\linewidth]{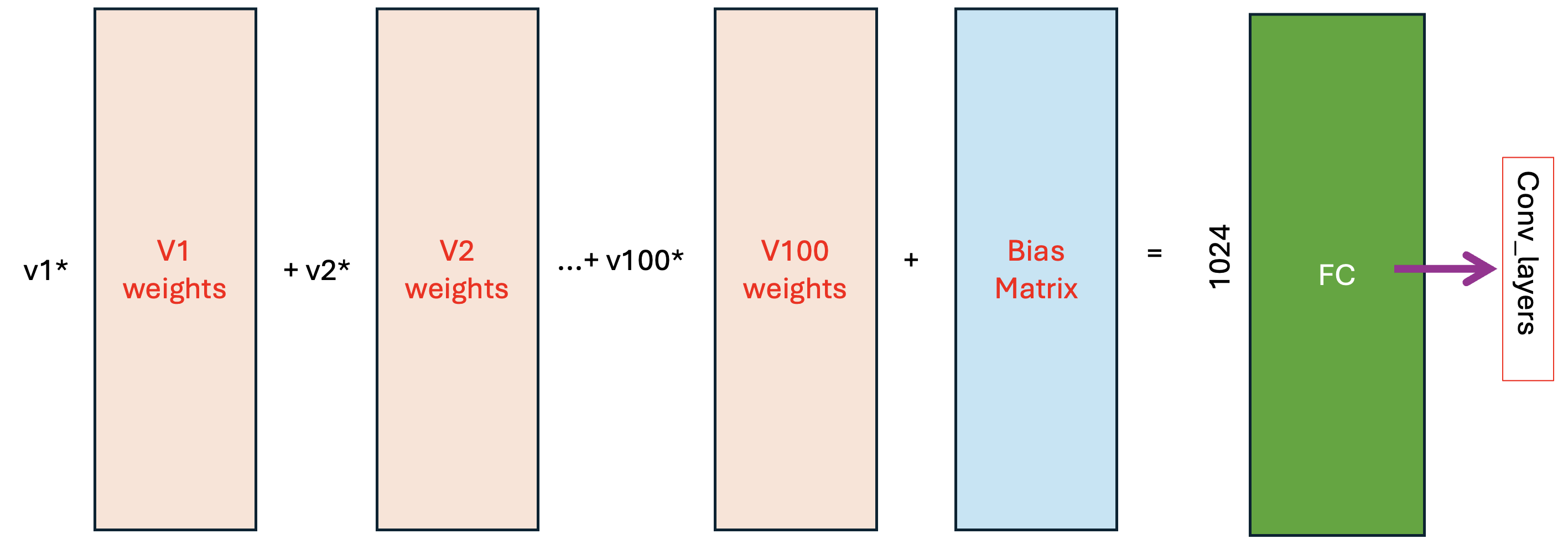}
    \caption{A schematic representation showing that the output of the FC-layer (before ReLU activation) can be represented as the sum of 100 weight matrices each scaled by a particular random variable value.}
    \label{fig:weight_matrices}
\end{figure}

In any latent space model, the compact vector of latent variables must be mapped to a higher dimensional space, and is typically accomplished via a fully-connected layer (e.g. \cite{belinkov2017analyzing}, \cite{ChworowskiRepresentation}). The fully-connected layer is thus the bottleneck through which latent variables affect subsequent layers. Critically, it is the only layer where there are trainable parameters unique to each latent variable (absent any residual or skip connections). In ciwGAN, the focus of the current paper, the fully connected layer projects a vector of 100 latent variable values to a vector of 16,384 values which is then reshaped to a $1024\times16$ feature-map representation. The output feature-map of the fully-connected layer, prior to any activation function, can intuitively be conceptualized as the addition of 100 weight matrices, each scaled by the value of the latent variable that the weight matrix corresponds to (Fig. \ref{fig:weight_matrices}). 

\section{Experiment 1: Comparison of latent variables by their weights}
This experiment demonstrates the utility of examining fully connected weights for latent space exploration. Prior work for latent space exploration in GANs, at least for linguistic data, has relied on labeling large samples of generator outputs according to particular linguistic features, and then testing which variables encode particular outputs by regressing samples of outputs against their corresponding latent space initializations (\cite{beguvs2020modeing}, \cite{chen2023exploring}). Here, we present an approach that supplements the aforementioned technique and which does not require labeling large quantities of generated outputs: we compare latent variables based on their trained weight matrices in the fully-connected layer. 

\subsection{Experiment 1a}
In this study we test whether the relative importance of latent variables can broadly be inferred from relative differences in the average magnitudes of their weights. As mentioned in Section 1.3, the output of the fully-connected layer, prior to any activation function, consists of 16,384 values, each of which is a weighted sum of all 100 latent variables (plus a trainable bias term for each output value). In GAN-based models like WaveGAN or ciwGAN, the latent variables are set to vary within a fixed absolute range during training. Because the weights multiply input values of the same scale, the relative magnitude of weights is directly interpretable. Moreover, variables can be compared based on the average magnitudes of their weight matrices: variables with higher average weight magnitudes are thus likely to have a greater impact on the waveform output overall. Comparing variables based on their average weight values thus offers a coarse-grained picture of which variables are likely to have a greater effect on the output overall. 

CiwGAN offers a testable prediction regarding which variables' weights should have higher magnitudes. In ciwGAN, the generator is optimized to maximize the mutual information between latent code inputs and the output to improve the Q-network's classification performance. It is thus possible that the latent codes' role in the training objective of ciwGAN should result in higher average weight magnitudes than random noise variables that are not directly implicated in facilitating the Q-network's classification task. For a ciwGAN trained with 9 latent codes corresponding to 9 lexical types in the training data, we test whether these latent codes have systematically different weight magnitudes than the random noise variables. 

Given that the FC-layer outputs are temporally structured feature maps, we also show how the latent codes and latent variables vary across the time-dimension by averaging across channels (e.g.~\cite{beguvs2022interpreting}). We train ciwGAN on audio samples where the lexical item occurs within the first half of the sample. Thus, if FC weight magnitudes reflect the importance of particular latent variables for modeling lexical items, weights within a particular latent code should be distributed across the time dimension such that higher weights are concentrated in regions of the channel that correspond to the regions of the outputs where lexical material is likely to occur. 

\begin{figure}
    \centering
    \includegraphics[width=0.75\linewidth]{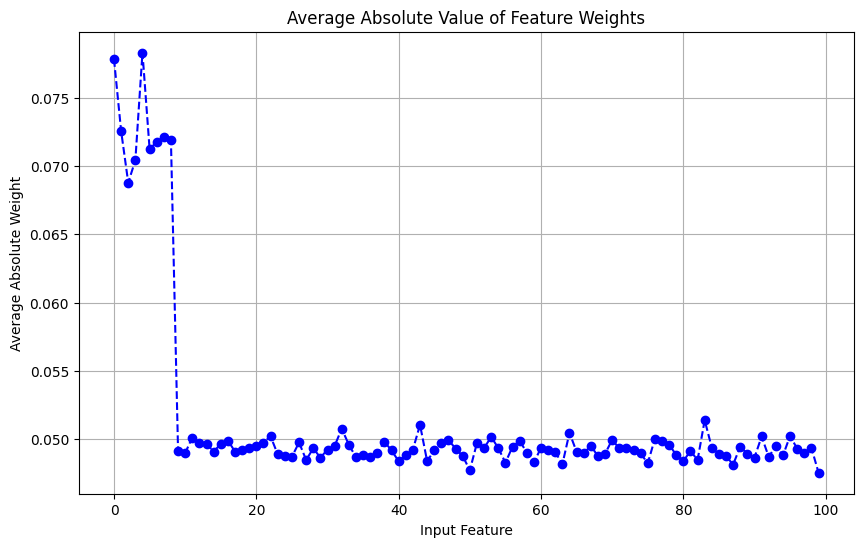}
    \caption{Average absolute weights of weight matrices for every variable in the latent space. The first 9 points are the latent codes (\textit{c}) while the rest of the points are the random noise variables (\textit{z}).}
    \label{fig:avg_points}
\end{figure}

\begin{figure}
    \centering
    \includegraphics[width=0.75\linewidth]{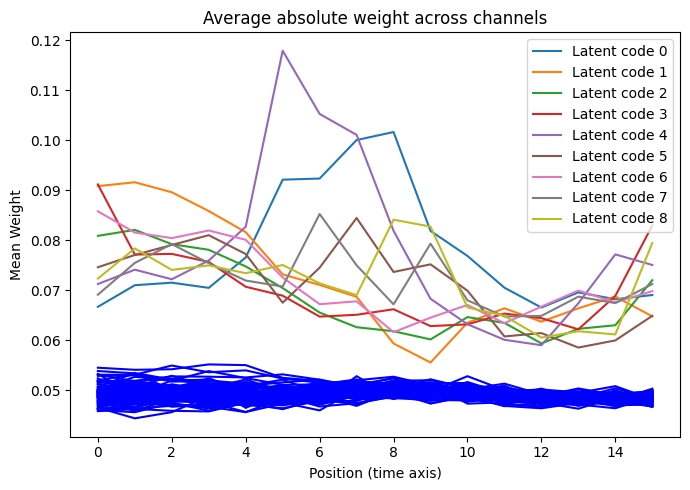}
    \caption{Average of absolute weight values along channel length. The blue curves represent the weights of uniformly distributed z-variables, while the colored curves represent latent codes. This illustrates that weights are concentrated in areas along the time-axis where the lexical item is situated in the training data.}
    \label{fig:avg_curves}
\end{figure}

\subsection{Results}
To test whether latent variables' importance is reflected in their weight magnitudes, we extracted the weight matrix of each latent variable, compute the absolute value of all 16,384 weight values, average these absolute values, and compare this average absolute weight for every input variable. As predicted, latent codes in ciwGAN exhibit greater average weights than the uniformly distributed z variables (see Fig. \ref{fig:avg_points}). This coarse-grained measure demonstrates that the relative importance of variables can be inferred from differences in their average absolute weights. In this case, the generator optimizes for the Q-network loss function by maximizing weights of the latent variables in the fully-connected layer. 

To visualize how weight magnitudes are temporally distributed across the channel of each variable-specific weight matrix, we plot the mean absolute value of weights within each of the 16 columns of 1024 values of the feature map (see Fig. \ref{fig:avg_curves}). We observe that weights are highest in magnitude near the linguistically relevant (non-silent) region of the channel which corresponds temporally to the beginning of the output signal. This is a consequence of the fact that the training data happened to be structured such that the word occupies approximately the first half of the 16,384-sample file. Thus, in this study, we demonstrates that even coarse-grained summary statistics of variable-specific weight matrices can offer insight into how the model allocates information between and within latent variables. As such, weight matrices can be an invaluable tool for preliminary bottom up exploration of the latent space, without the need for generating and labeling large quantities of waveform outputs. 

\subsection{Study 1b: weights as inputs into convolutional layers}
Given that each variable's weight matrix in the fully-connected layer shares the feature map structure of the output of the entire Fully-connected layer itself, these weight matrices can be directly passed into the convolutional layers to produce a waveform output. In other words, learned weights can be treated as inputs into the convolutional block of the model because their feature-map configurations directly correspond to patterns of activation for any FC layer output sampled "naturally" from the latent space. To our knowledge, no prior work has used weight matrices as inputs into convolutional layers. 

For each given latent variable, we test whether linguistically interpretable information is evident from generating an output from that latent variable's weight matrix. Given the success of lexical learning, namely the consistent one-to-one mapping between latent code and lexical item, we expect that each latent code's weight matrix should yield a waveform output that corresponds to a lexical item in the training data, when passed as a feature map into subsequent convolutional layers.\footnote{The current model does not include an activation function after the fully-connected layer. Prior work has used a ReLU activation. The same approach for such models should first apply the ReLU activation to the weights.}

\subsection{Results}
\begin{figure}
    \centering
    \includegraphics[width=1\linewidth]{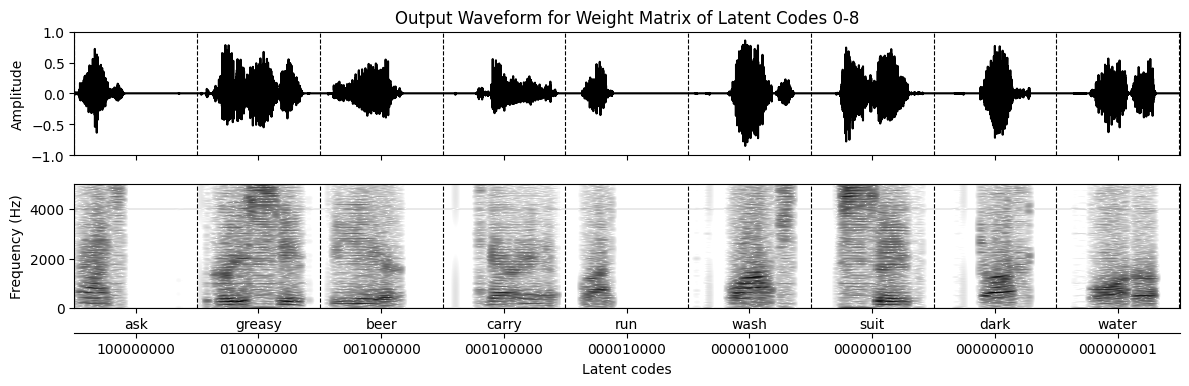}
    \caption{Output waveforms, each derived from passing variable-specific weight matrices as inputs into the convolutional layers. Each waveform unambiguously matches one of the nine lexical items that the network was trained on. For compactness, only the first 11000 samples of each 16340-sample output is shown.}
    \label{fig:code_outputs}
\end{figure}

For the ciwGAN trained on 9 latent codes, inputting each code's $1024\times16$ weight matrix yields a discernible waveform output corresponding uniquely to one of the 9 lexical items in training (Fig.~\ref{fig:code_outputs}). The spectrograms in Fig.~\ref{fig:code_outputs} show a clear differentiation for each latent code with acoustic structure corresponding to our manual transcription. 

\begin{figure}
    \centering
    \includegraphics[width=1\linewidth]{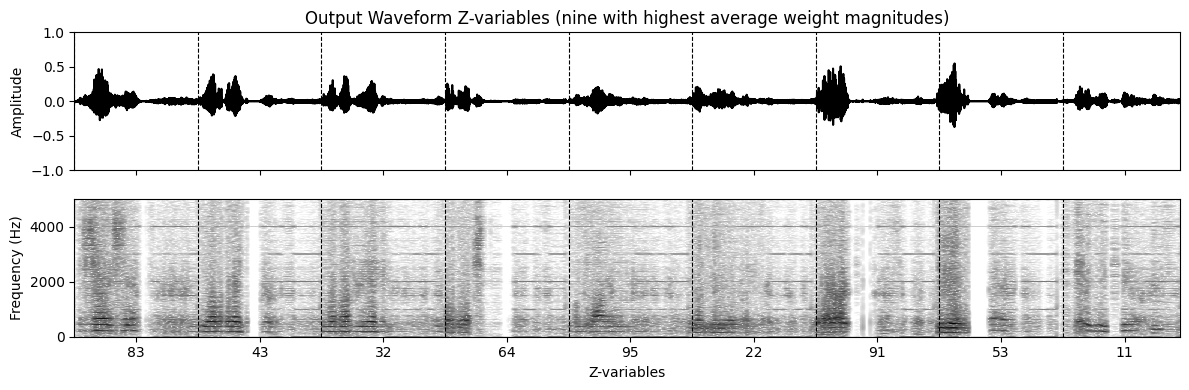}
    \caption{Output waveforms derived from passing variable-specific weight matrices. Weight matrices for uniformly distributed variables do not appear to encode specific linguistic structures. }
    \label{fig:z_variables}
\end{figure}

As a comparison, we pass uniformly distributed variables ($z$) to test whether variables are likely to encode specific lexical or sublexical information. We do not make any specific predictions regarding the outputs of any \textit{z}-variables. In Figure \ref{fig:z_variables}, we show the 9 \textit{z}-variables with the highest average weight magnitudes. It should be noted that in the course of training, \textit{z}-variables' weights are never passed into the convolutional layer independently of latent codes, or independently of the bias matrix, and therefore their outputs should be interpreted as mere approximations of the likely effect they have when added to latent codes. The spectrograms of the z-variable weight matrices do not show lexical information, unlike those of the latent code weight matrices. 

\subsection{Discussion}
In this study, we introduced and validated two ways of using the Fully-connected layer for bottom-up latent-space exploration. We show that variables which we a priori expect to be more important given the training objective of ciwGAN, end up with higher average weight magnitudes over the course of training. We also show that using variable-specific weights as inputs into subsequent layers provides a waveform representation of what linguistic/acoustic quality that variable encodes. Both approaches for bottom-up latent space exploration are applicable to any framework in which a latent space is projected into a higher dimensional space via a fully-connected layer. 

\section{Experiment 2: Testing for lexically-independent vowel representations in the Fully-connected layer}

In Experiment 1, we observe that as a consequence of the model successfully accomplishing the training objective of ciwGAN, the weight matrix of each latent code has greater average weight magnitudes than all \textit{z}-variables, and when passed as the input into the convolutional block, each code generates a unique lexical item from training. This result supports prior work (\cite{beguvs2021ciwgan}) demonstrating the success of ciwGAN for lexical learning by demonstrating a one-to-one mapping between lexical type in training and latent code. \cite{begus22Interspeech} also demonstrate that the models learn sublexical structure in the fiwGAN architecture. However, it is unclear whether the current ciwGAN model encodes correspondences between lexical items in a meaningful way in the FC layer. One way of interpreting the fact that the latent codes uniquely map to lexical items is that the model represents sublexical information in entirely non-compositional and lexically-specific way. Under this scenario, ostensible sublexical units are so lexically-specific that acoustic similarity of sublexical units should not be reflected in their representational similarity in the weights of the fully-connected layer. Alternatively, it may be the case that similar sublexical units are encoded similarity in the fully connected layer. For example, is the category /i/ encoded similarly in the weights of "greasy" as it is in the weights of "beer"? 

\begin{figure}
    \centering
    \includegraphics[width=0.5\linewidth]{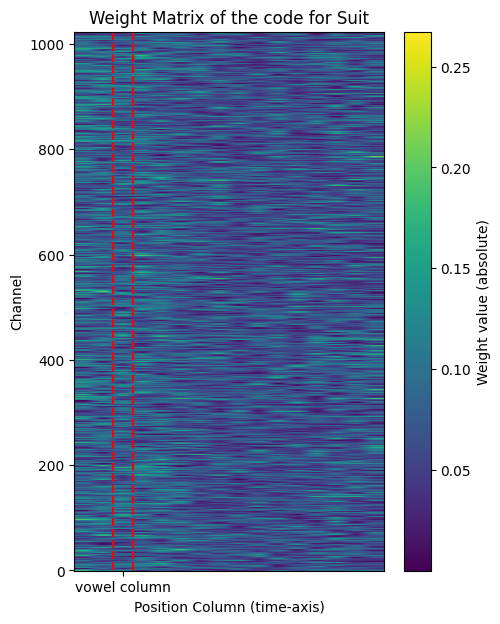}
    \caption{Visualization of entire weight matrix for the latent code that encodes "suit".  The values within the red dashed lines represent the vowel column for the vowel in the word "suit". We chose the most appropriate column to extract by testing which column yields only vowel information when input into the convolutional layers. We extract such columns for weight matrices of all 9 latent codes. }
    \label{fig:vowel_column_featuremap}
\end{figure}

The weights of the fully-connected layer allow for testing for evidence of principled sub-lexical learning in ciwGAN. In order to test whether the model learns similar representations for similar sublexical sounds across distinct latent codes, we leverage the temporal feature map structure of the variable-specific weight matrices, to pinpoint the column in the weight matrix that best corresponds to a particular vowel in a particular lexical item. Fig.~\ref{fig:vowel_column_featuremap} shows the entire feature map for the word ``suit'' and the weight column that best corresponds to the vowel in "suit" is marked between two red lines. 

\begin{figure}
    \centering
    \includegraphics[width=0.5\linewidth]{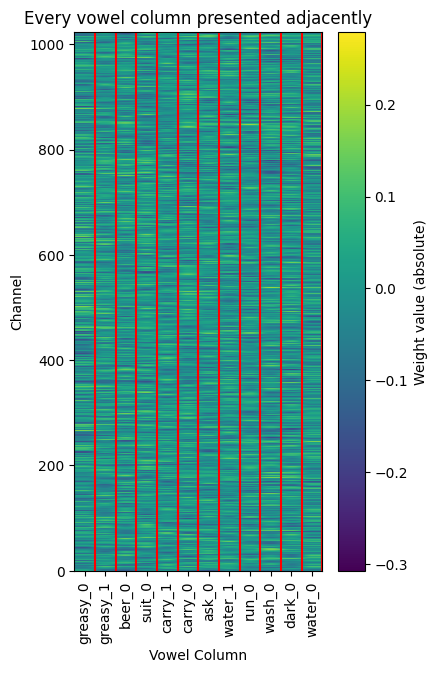}
    \caption{Weight columns for every vowel. These 12 columns are extracted from the latent code weight matrices. Each column represents a raw input into the convolutional layers and each is used in computing correlations between vowel representations. ``\_0'' denotes the first vowel in the word, while ``\_1'' denotes the second vowel in the word.}
    \label{fig:twelve_columns}
\end{figure}

\begin{figure}
    \centering
    \includegraphics[width=1\linewidth]{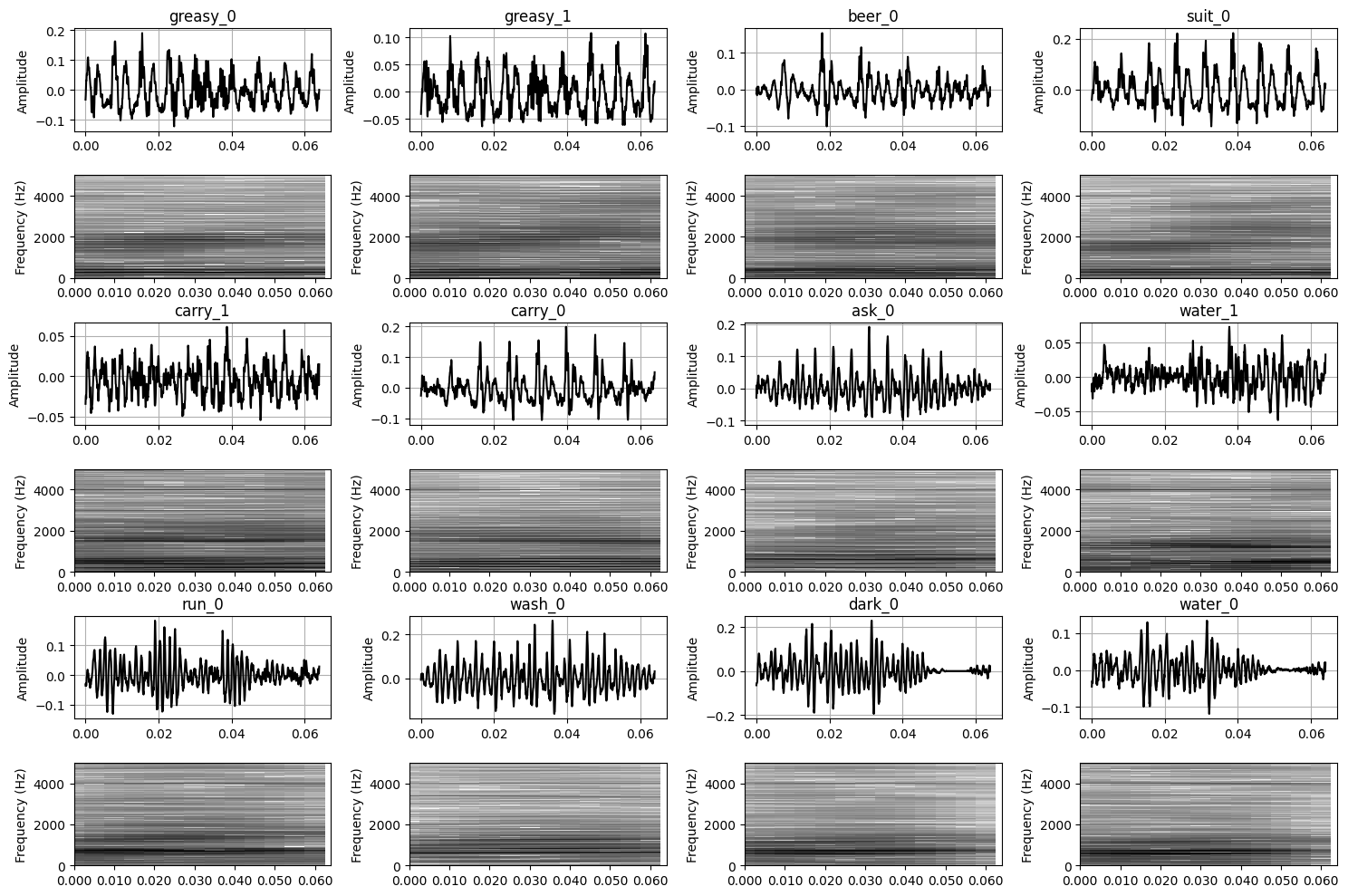}
    \caption{The waveform output of each isolated vowel column when passed as input into the convolutional layers. The derived spectrogram is plotted below each waveform.}
    \label{fig:vowel_column_waves}
\end{figure}

To determine the column that best corresponds to a particular vowel, we pass individual feature columns of the latent codes' weight matrices and select the column that contains only vowel information in its output (Fig.~\ref{fig:vowel_column_waves}). We extract a total of 12 weight columns from the 9 lexical items' weight matrices: 6 vowels from words, ``greasy'', ``carry'' and ``water'' that have two syllables, and one vowel from each of the six monosyllabic words. The weight values of these 12 vowel-columns are shown in Fig.~\ref{fig:twelve_columns}. This experiment thus also shows that subselecting a column in the FC layer can result in a single vocalic segment in the output, which is useful for isolating how the network represents particular sublexical categories. 


\begin{figure}
    \centering
    \includegraphics[width=1\linewidth]{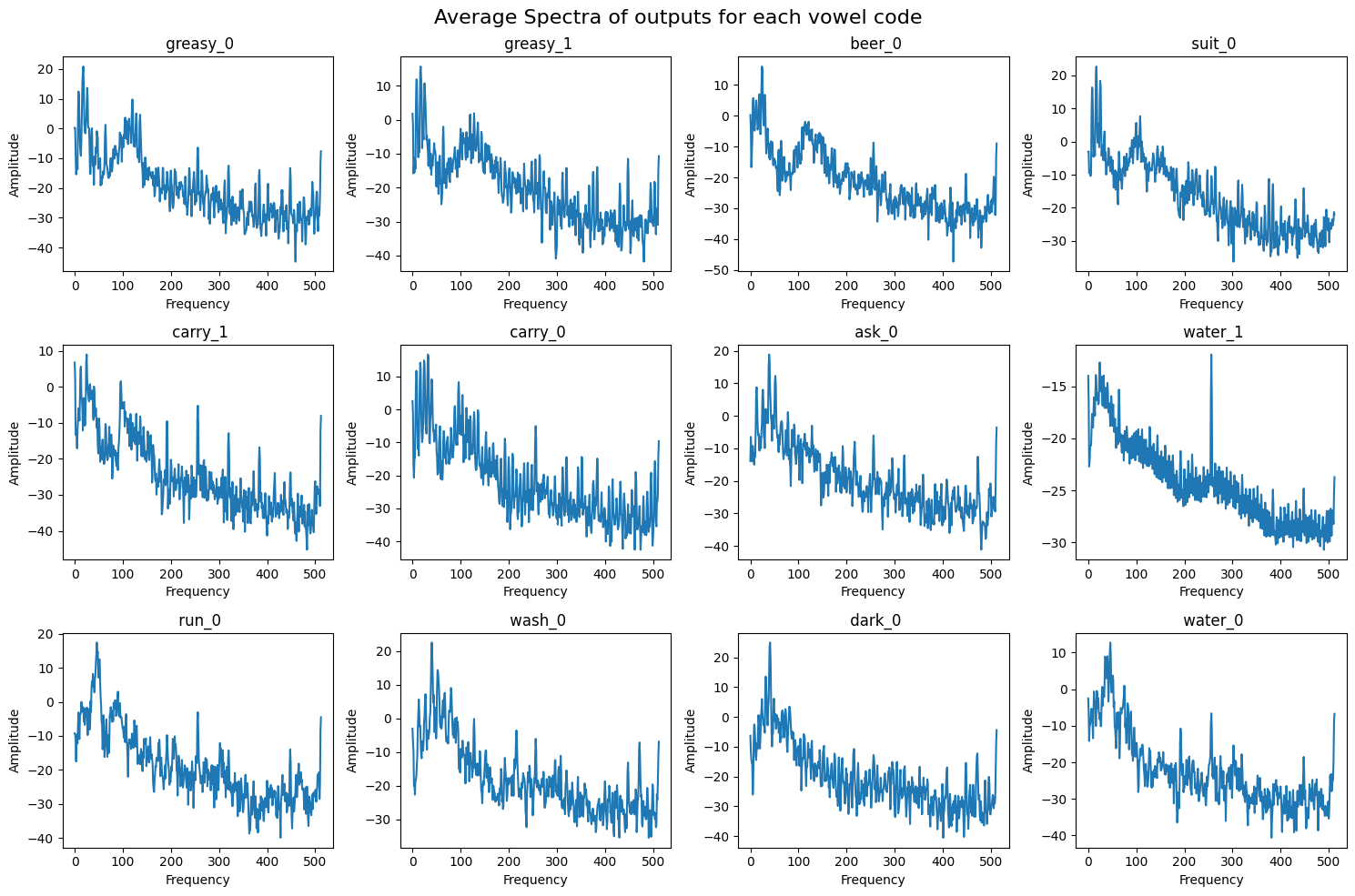}
    \caption{Averaged spectra for the output of each vowel code. High-front vowels have a bimodal peak, representing a high F1 and low F2, while mid-low vowels have a single peak, a reflection of a higher F2 and lower F1. These represent the input into the correlation matrix in Fig.~\ref{fig:spectra_correlations}}
    \label{fig:averaged_spectra}
\end{figure}

We then compare these 12 vowel-columns as well as their corresponding outputs.  Specifically, we compute a correlation matrix to test whether phonetically similar vowel categories have FC-codes that are more correlated with each other than those of phonetically distinct vowel categories. For example, if acoustic similarity is reflected in representational similarity, the weight-column for the first /i/ in ``greasy'' should be more correlated to the weight column for the vowel /i/ in ``beer'' than for the weight column corresponding to the /a/ vowel in ``wash.'' We compute a $12\times12$ correlation matrix for weight columns for all vowels. To assess whether acoustic similarity reflects representational similarity of the weight matrices, we compute the same correlation matrix for the corresponding outputs of these weight columns when passed through the convolutional layers. We derive a spectrogram for each waveform output and then average the amplitude values across the duration of the vowel to get a 1-dimensional average spectrum, consisting of 1000 values (Fig. \ref{fig:averaged_spectra}). If representational similarity in fully-connected weights translates into acoustic similarity, the correlation matrix of FC weights and those of the respective vowel spectra should be similar to the correlation matrix of averaged spectra of the vowels. 

\subsection{Results}
\begin{figure}
    \centering
    \includegraphics[width=0.5\linewidth]{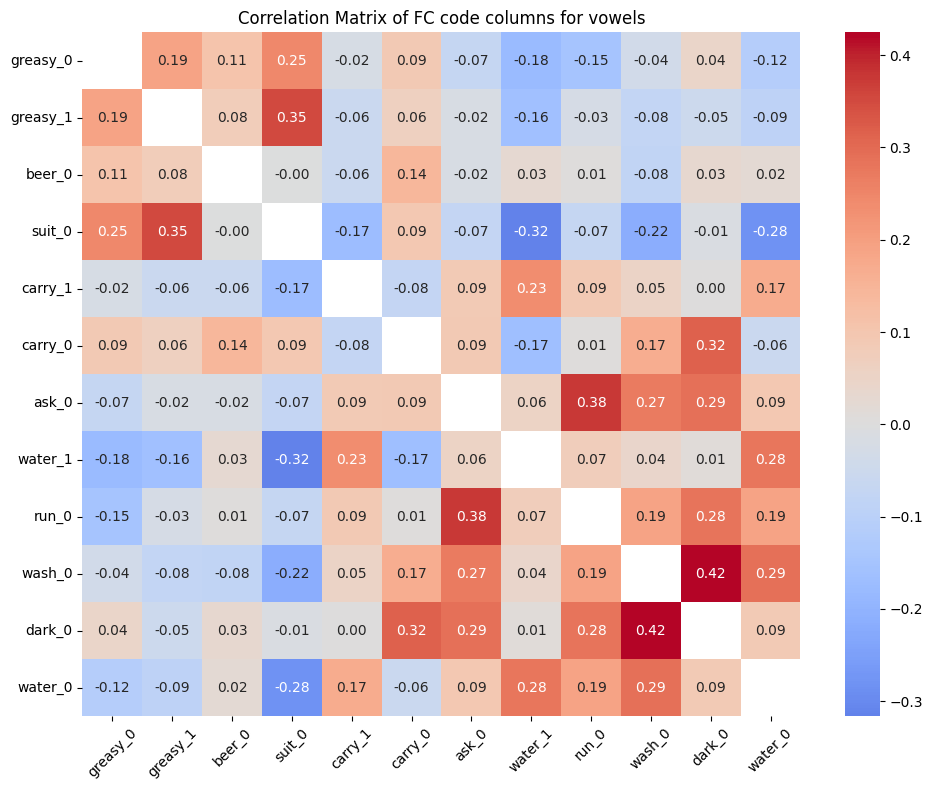}
    \caption{Correlation Matrix of feature map FC-codes (columns in the FC feature map) that produce each vowel category. Red represents positive correlations while blue represents negative correlations. There is evidence that with the exception of /i/ in ``carry'', FC codes for high vowels are positively correlated with each other and not with codes for non-high vowels and vice versa.}
    \label{fig:code_correlations}
\end{figure}

\begin{figure}
    \centering
    \includegraphics[width=0.5\linewidth]{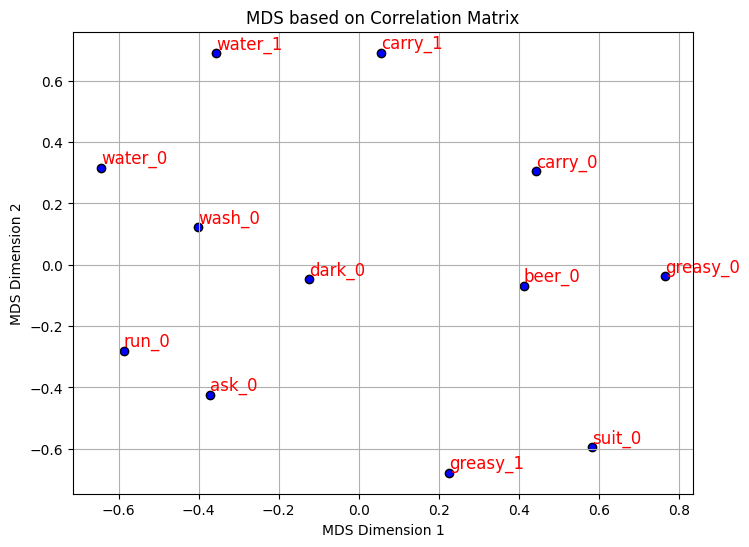}
    \caption{Multidimensional-scaling plot. The MDS algorithm infers a 2d space on which to situate the points based on pairwise distances from the correlation matrix. Higher and fronter vowels are grouped to the right (positive values of dimension 1), while lower, backer or ambiguous vowels tend to be grouped to the left. This shows that the representational encoding of the vowels in the fully-connected layer broadly corresponds to their phonological vowel quality differences.}
    \label{fig:mds_code_correlations}
\end{figure}

The correlation matrix of FC codes shows that codes for high vowels tend to be more correlated with each other than with codes for low vowels and vice versa (Fig.\ref{fig:code_correlations}). We visualize between-code distances using a multidimensional scaling algorithm, which takes pairwise distances as its input (1 - correlation value), and calculates the most likely configuration of individual points on a 2d space based on the pairwise distances. This plot confirms that phonologically similar vowels share similar encodings: the vowel columns of all unambiguously high vowels occur to the right along dimension 1, while all  phonologically low vowels occur to the left (Fig.\ref{fig:mds_code_correlations}). Thus, there is evidence for the encoding of word-invariant vowel differences in Fully-connected layer. For example, the feature-column determining the /i/-like sound in ``greasy'' is correlated with that determining the /i/ in "beer" and is negatively correlated with the column determining the /a/ sound in ``wash'', ``dark'', and ``water.'' 


\begin{figure}
    \centering
    \includegraphics[width=0.5\linewidth]{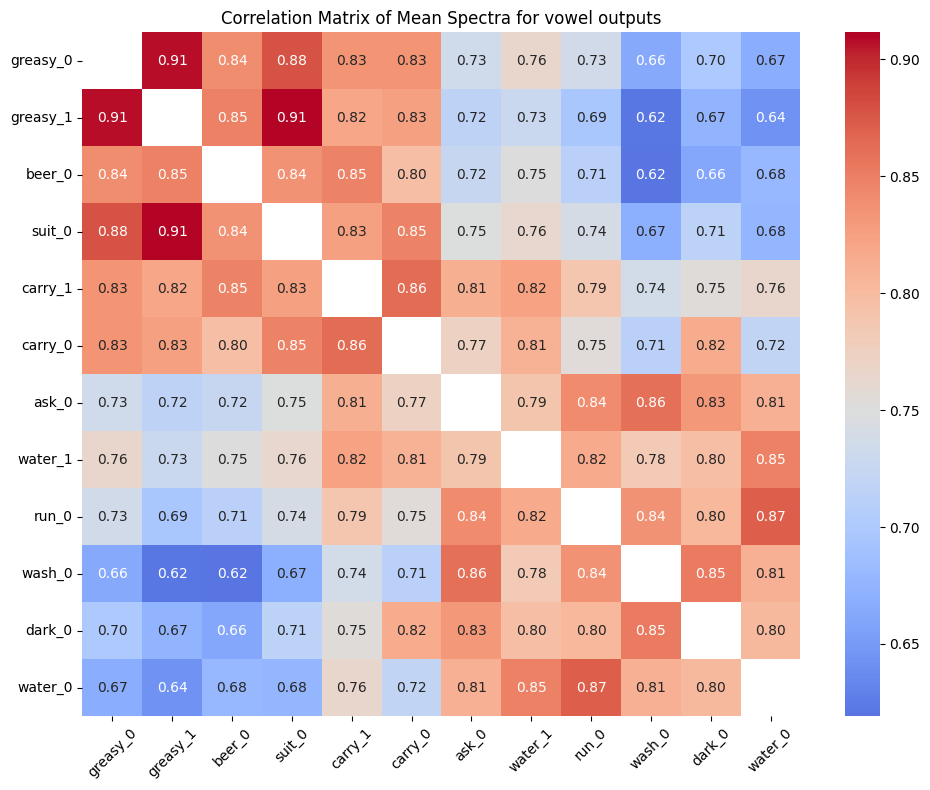}
    \caption{Correlation Matrix of spectra (spectrograms averaged across the time dimension) derived from the waveform outputs of each variable's weight column. Warmer colors represent greater similarity, while cooler colors represent a smaller degree of similarity. Given that all correlation values are positive, the red cells are above-average correlations, and blue cells show below-average correlations.}
    \label{fig:spectra_correlations}
\end{figure}

The correlation matrix of the output spectra provides further verification to ensure that acoustic similarity between sounds broadly corresponds to similarity in the weight columns that determine these vowel sounds (Fig. \ref{fig:spectra_correlations}). The relative deviation of the column for the /i/ in "carry'' from the rest of the high-vowel columns is also reflected in the spectral information of the output where it exhibits relatively higher acoustic correlations with the low vowels than the rest of the high vowels. 


Taken together, these data validate the notion that similar-sounding vowels are encoded in similar ways in the FC layer, both within and across weight matrices of lexically-specific latent codes. Crucially, this representational economy is emergent and is neither an explicit part of the network's training objective nor an attribute of the training data that consists of 9 lexical types that be readily mapped to 9 latent codes. In fact, the generator's task of optimizing the Q-network's performance could exert a pressure \textit{against} overlapping representations in favor maximally distinct representations for the different lexical items, even if these lexical items share sublexical similarities. While previous results have shown evidence that phonemic categories like /s/ map uniquely to latent variables (\cite{begus22Interspeech}), in this model, where latent variables are optimized to capture entire lexical items, stable sublexical structure is still evident in activations of the fully connected layer.  

\subsection{Experiment 2b}
In Experiment 2a, we showed that vowel-inducing columns in the fully-connected layer can be used to produce single vowels when passed to the convolutional layers, and we showed that similar vowels for distinct lexical codes are encoded similarly in single vowel-inducing weight column. In this experiment, we explore the extent to which these feature columns can function as compositional units. More specifically, we explore the robustness with which these weight columns yield particular linguistically interpretable input in the output when they are passed to the convolutional layers alongside other columns. This study is exploratory, and assesses the consequences of recombining weight columns into feature-map configurations that are impossible to derive from latent space manipulation alone. 

First, we test the consequences of passing a feature map that comprises adjacent copies of the same feature column to the convolutional layers. When the feature-column for the /i/ in ``greasy'' is copied three times along the time dimension and passed to the convolutional layer as a $1024\times3$ matrix, we observe that the resulting output waveform contains an /s/ sound at the beginning of the output waveform. 

\begin{figure}
    \centering
    \includegraphics[width=0.5\linewidth]{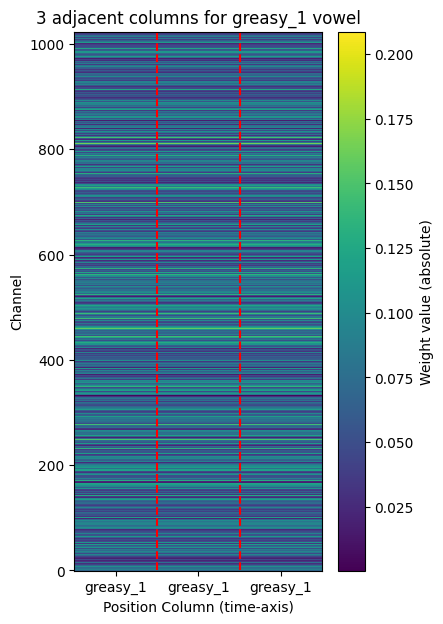}
    \caption{Visualization of the manually-constructed input into the convolutional layers we use to test how the outputs weight columns that produce vowels in isolation changes in the context of other weight columns.}
    \label{fig:identical_adjacent}
\end{figure}

\begin{figure}
    \centering
    \includegraphics[width=0.5\linewidth]{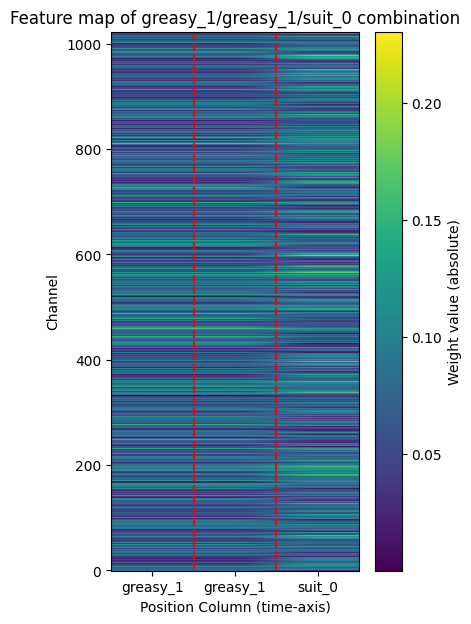}
    \caption{Visualization of one example of a manually-constructed input into the convolutional layers we use to test how feature-maps of mismatched weight columns columns yield predictable outputs.}
    \label{fig:different_adjacent}
\end{figure}

The same pattern in the output emerges when the input to the convolutional layers consists of a weight column for the vowel in ``suit'' and is copied three times as a $1024 \times 3$ matrix (see Fig.~\ref{fig:identical_adjacent}). We then produce outputs where the feature column for the vowel in ``suit'' is situated alongside a pair of feature-columns for the second vowel in ``greasy'' in order to test whether the appearance of an /s/ before the vowel occurs when feature columns from different weight matrices are included in the same input feature map. In other words, we test whether feature columns behave like compositional units. Given that the configuration of weight columns into the feature map suit0+suit0+suit0 yields an /s/ followed by a vowel, we should expect the same pattern to emerge in any combination of suit0 and greasy1, such as greasy1+greasy1+suit0 (see Fig. \ref{fig:different_adjacent}) 
\begin{figure}
    \centering
    \includegraphics[width=0.5\linewidth]{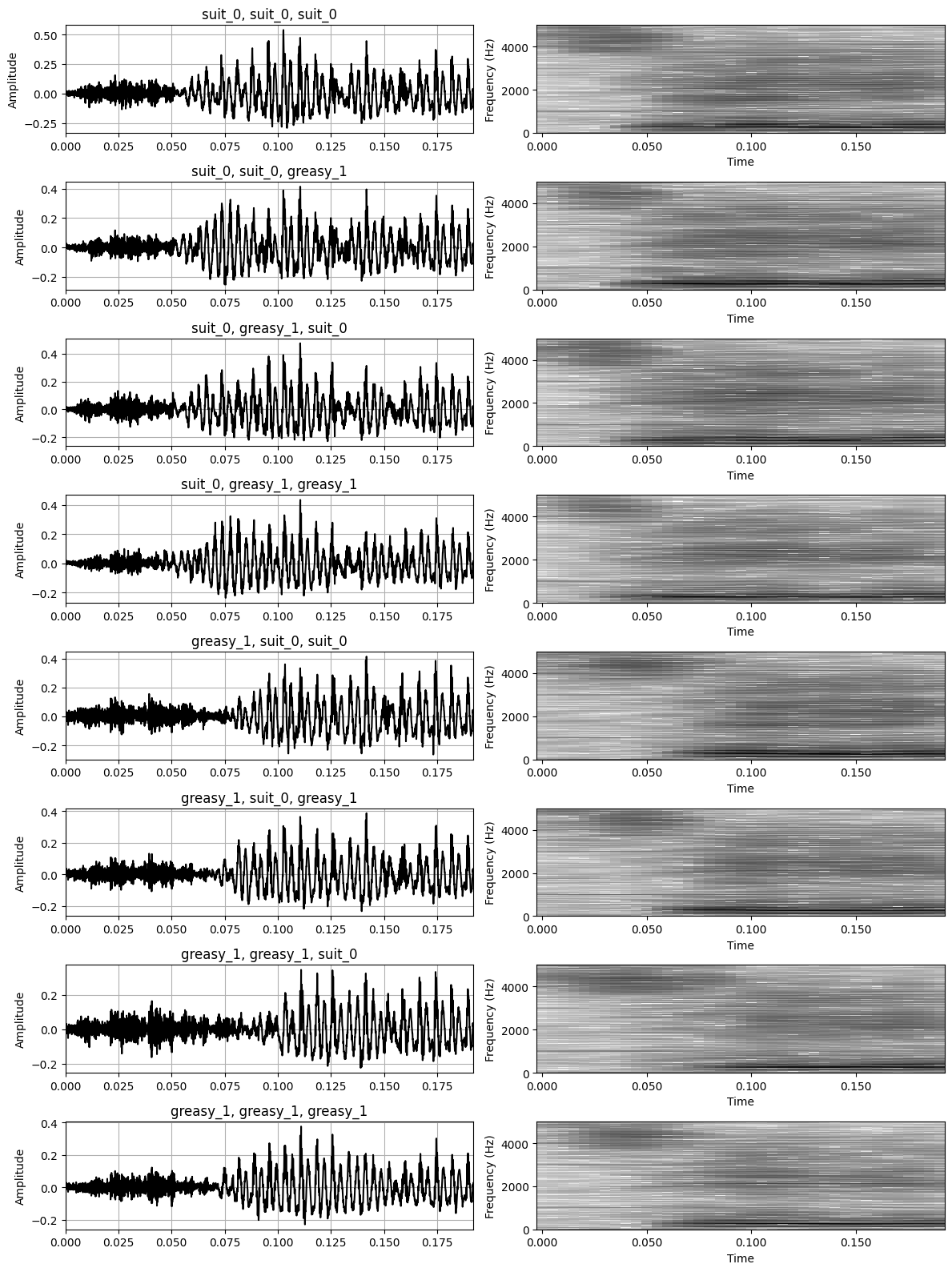}
    \caption{Outputs for every 3-column feature map consisting of columns that in isolation encode the first vowel in ``suit'' or the second vowel in ``greasy''. All outputs resemble a /si/- or /su/-like sound. This demonstrates that the feature columns can themselves function as compositional units in that their interaction with other weight columns in manually assembled 3-column feature maps is consistent and predictable at a high level. Only the relative amount of frication versus vowel information varies across these feature map. Waveforms and corresponding spectrograms are presented in separate columns for ease of comparison.}
    \label{fig:all_adjacent_outputs}
\end{figure}

\subsection{Results}
The waveform outputs of all 8 feature maps contain an /s/ followed by vowel information (Fig. \ref{fig:all_adjacent_outputs}). This observation provides evidence that temporally-arranged feature columns in the fully-connected weights exhibit properties of compositional units in that they exhibit roughly the same non-linear contextual effects in outputs when they are situated alongside each other. Thus, these columns of learned weights are sufficiently representationally abstract that they can, in principle, function as interchangeable compositional units in the sense that they produce predictable contextual effects. 

That this configuration of feature columns produces an /s/ before the vowel in all cases indicates that aspects of both the vowel and consonant are encoded in the weight columns that produces an isolated vowel when passed to the convolutional layers in isolation. One explanation for this output pattern is that some of the channels or combinations of channels in the feature-column exclusively encode an /s/ but are not expressed when the column is passed to convolutional layers in isolation. Alternatively, it may be the case that the emergence of /s/ in the presence of another feature column is a consistent property of features that encode these lexical items. 

To assess how /s/-triggering and vowel-triggering information is distributed across channels in the feature column for \textit{greasy\_1} and \textit{suit\_0}, we test whether this property emerges when smaller sub-samples of channels are passed to the convolutional layer. Specifically, we take 16 samples of 64 channels at regular intervals (e.g. 1-65, 65-128 etc.), and assign zero to the rest of the channels in the 3-column feature map, effectively neutralizing the activations of the other channels. Isolating the effect of certain values in a feature map by assigning zero to others has been previously used in research on computer vision (\cite{zeiler2013visualizingunderstandingconvolutionalnetworks} but, to our knowledge, we are the first to use to evaluate linguistic representations in a speech-generation model. 

If a given feature column contains both activations for the vowel and activations for /s/, we should observe the degree of /s/ and the degree of vowel information vary by sample. If the emergence of /s/ is instead a deterministic property of vowel-specific activations occurring simultaneously in the feature map and their interaction with convolutional kernels, we should observe a consistent pattern of both /s/ and vowel information across any given output. 

\begin{figure}
    \centering
    \includegraphics[width=0.75\linewidth]{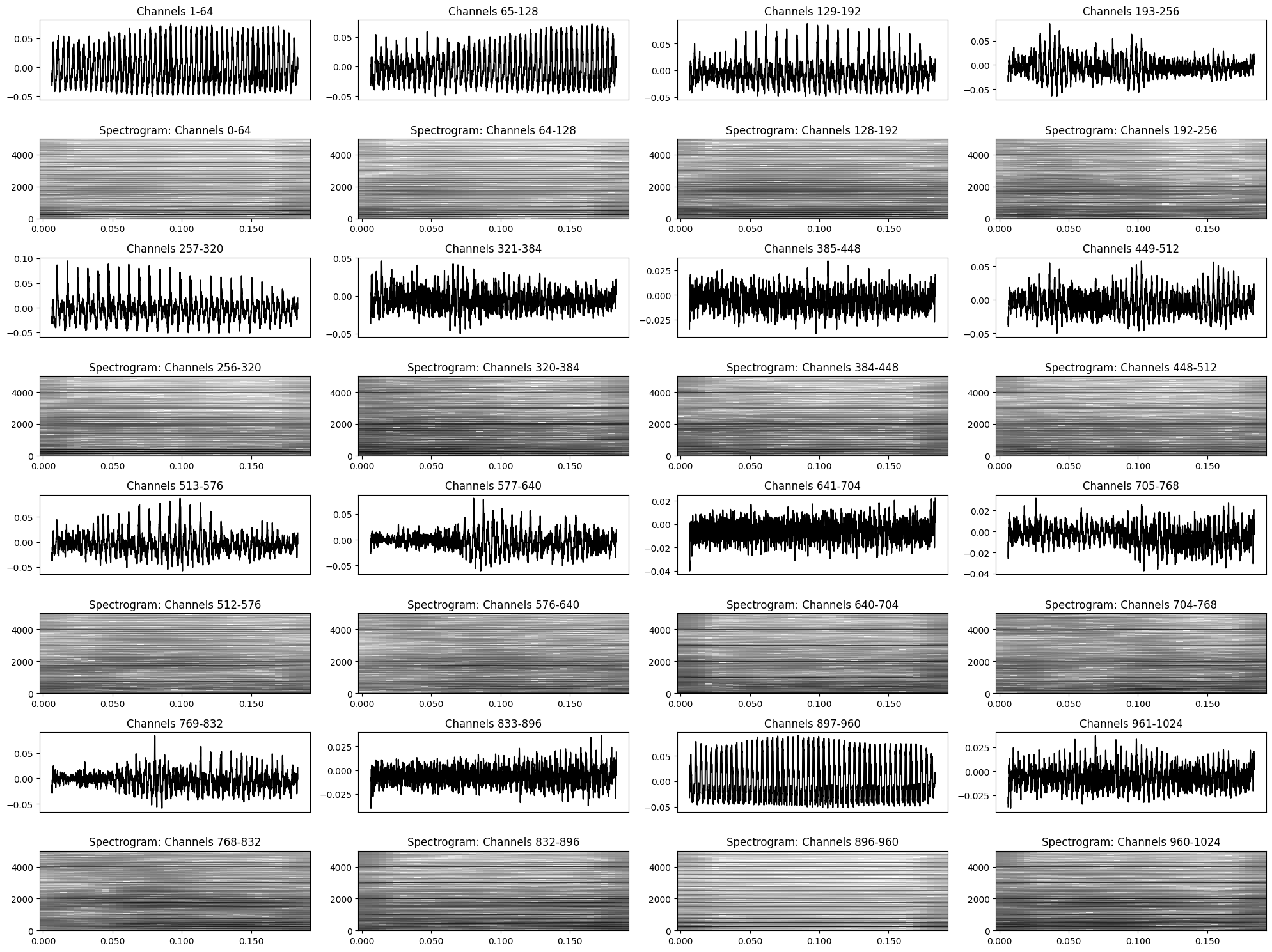}
    \caption{Waveforms generated by passing 64-channel samples of the full weight matrix consisting of 3 identical columns specifying the second /i/ in ``greasy''. The fact that the outputs of some of these samples homogeneously contain only vowel information while others contain homogeneously contain only frication noise illustrates how activation for different kinds of output information is distributed across different channels in the feature column. These vowel outputs tend to resemble high vowels in their formant structure. }
    \label{fig:channel_subsets}
\end{figure}

We find that the 64-channel samples vary according to the relative amount of vowel and /s/ information they encode (Fig. \ref{fig:channel_subsets}). For example, Channels 577-640 mirror the output of the entire weight column. Channels 641-70 contain just frication noise, and the outputs of channels 257-320 contain only vowel information. Thus, when considering these subgroups of neurons (pixels) as units of linguistic representation, vowel information is still representationally dissociable from non-vowel information. The fact that /s/ information is more likely to surface early in the input is still likely a consequence of how the convolutional kernels interact with these columns. The variability of waveform outputs in Fig.~\ref{fig:all_adjacent_outputs} is thus a function of which channels in each permutation of weight columns happens to align in such a way to yield more /s/ information or more vowel information. 

\subsection{Discussion}
In summary, our exploration of the fully connected weights of ciwGAN's generator yields evidence that the generator learns a representationally consistent way of representing the acoustic difference between high and low vowels across different lexical items. This helps address a possible concern of ciwGAN that lexical items are encoded wholistically in uninterpretable ways, and that encodings of lexical items need not correspond to each other in linguistically meaningful ways. We show evidence of the opposite: similar vowels are encoded in similar ways across different lexical items. 

We also observe that the phonetic manifestation of weight columns that encode vowel information varies by context: weight columns that produce an /i/ in isolation also produce an /s/ when placed adjacently to other such columns or copies of themselves. Thus, while columns in the feature map output of the FC layer are methodologically useful for finding evidence of abstract-like representations that can in principle be compositional, there is no one-to-one mapping between these columns and phonological categories in the output. However, these contextually dependent effects are nevertheless predictable and occur in lexically-independent ways. Specifically, the fact that a vowel-specifying column can produce an /s/ when placed alongside a copy of itself reflects both the architectural fact that temporal units of the FC-layer overlap in their effects on the output. Methodologically, this is the first study, to our knowledge, to isolate and recombine feature map columns in the fully connected layer and pass them to the convolutional layers to produce an output. We also demonstrate that  feature maps with subsets of channels assigned zero yield interpretable outputs when input into the convolutional layers. Manually assigning zero to particular channels in a feature map is a flexible and potentially invaluable tool for isolating which information is encoded in which channels. 

\section{General Discussion}
Recent approaches to modeling speech using GANs have generally explored the mapping of linguistic variables to the latent space by labeling large amounts of generated outputs and exploring correspondences between such labels and latent variables (e.g. \cite{beguvs2020modeing}, \cite{begusCDEV}, \cite{beguvs2021ciwgan}, \cite{barman2024unsupervised}, \cite{chen2023exploring}).  In this study, we demonstrate the benefit of considering latent variables in terms of their variable-specific weights for model interpretability. In the FC-layer, we find evidence for sub-lexical structure that is not readily-mapped to particular latent variables, that lends itself to compositional recombination. 

\subsection{Representations of Phonemes in GANs and Humans}
Our investigation of the fully-connected weights in a ciwGAN shows that the fully-connected layer encodes sublexical structures shared across words in similar ways. Thus, lexically invariant ways of encoding vowel information emerge in the fully-connected layer. We additionally show that selecting the vocalic parts of the FC layer allows generating individual vowels (such as [i]).  While the vowel-specific representations in the FC-layer are categorically stable, they are nevertheless contextually conditioned and contain activations that correspond to neighboring sounds, as evidenced by the fact that the second vowel in "greasy" and the column for the first vowel in "suit" contain activations for a preceding /s/ that do not manifest when that column is passed to the convolutional layers in isolation.  The fact that these representations are encoded in context in predictable way is broadly consistent with theories such as exemplar theory that propose the existence of phonetically and contextually rich sublexical representations (\cite{johnson1997speech}, \cite{pierrehumbert2001exemplar}, \cite{goldrick2023advancement}). More generally, these findings are consistent with the notion that the acoustic signal contains sufficient cues to for the learner to extract stable sub-phonemic categories (e.g. \cite{shain-elsner-2020-acquiring}, \cite{begus22Interspeech}, \cite{abdullah2024representation}). 

\subsection{Sampling from values of internal states}
Among the key innovations of this paper is that rather than sampling outputs from the latent space, we manipulate the representational blueprints of latent variables directly. This approach can be characterized as one way of sampling from the multidimensional space of possible feature-map inputs as opposed to simply exploring the latent space. Considering representational configurations that do not occur in training and that are not part of the configurations encoded in weight matrices is a potentially invaluable tool for model interpretability and one which is more flexible than latent space manipulation. Among other things, it can be used to probe the abstractness of representations and the nature of apparent rule like behavior detected by latent space manipulation (\cite{beguvs2020modeing}, \cite{beguvs2021identity}), as well as probe the division of labor between the fully-connected layer and subsequent layers in capturing particular output phenomena. 

We also demonstrate that sampling outputs from subsets of the fully-connected layer can yield insight into what the model learns, which would be unavailable via latent space interpolation. For example, through sampling the outputs of particular channels in the FC layer, we also showed that there is a large degree of redundancy in how linguistic information is encoded in the network across channels. This suggests that successful learning may occur even when the model is trained with fewer parameters. Some work has tested the effect of shrinking or expanding the latent space bottleneck in auto-encoders for image generation (e.g. \cite{manakov2019walking}). Similar work should explore whether shrinking the number of parameters of ciwGAN or similar latent space models is costly or beneficial to the model's ability to form abstract-like linguistic generalizations. 

The proposed interpretability techniques on the FC layer can also serve to as the basis for further comparison between brain responses to human speech and artificial neural network responses to the same signal. Prior work has shown that interpretability techniques on  convolutional layers \cite{begusZhouIEEE} allow for a  raw, untransformed comparison between the complex auditory brainstem signal (cABR) and the shallower convolutional layer (\cite{begusZhouZhao}). Our current work allows the comparison between the deepest layer in the convolutional neural network paradigm and the human brain.

\section{Conclusion}
The current study demonstrates the methodological utility of manipulating the fully-connected layer as a tool for exploring the latent space. We propose two techniques for interpreting the linguistic effects of the FC layer and apply them to the ciwGAN architecture in Experiment 1 and 2. We show that lexically-invariant sublexical structure that is not necessarily mapped uniquely to a latent variable emerges in the trained weights of the dense layer. By analyzing between- and within-variable variability of variable-specific weights in the fully-connected layer, we show that learning of stable sublexical categories can emerge in ciwGAN. This shows that the model does not merely ``hard-code'' lexical items wholistically, but reuses similar representations to represent similar sounds. While prior work has demonstrated representational similarity of similar linguistic units in DNNs, we go a step further by splicing and substituting representational units in feature maps and showing to what extent they are in fact interchangeable.  Future work should extend these techniques for model evaluation in other frameworks that rely on latent spaces, such as auto-encoders. 

\appendix

\bibliographystyle{unsrt}

\end{document}